\documentclass[11pt]{article}
\usepackage{geometry}
\usepackage{coling2020}
\usepackage{times}
\usepackage{url}
\usepackage{latexsym}
\usepackage{microtype}
\usepackage{tabularx}
\usepackage{graphicx}
\usepackage{adjustbox}
\usepackage{amsmath}
\usepackage{dirtytalk}
\usepackage{enumitem}
\usepackage{booktabs}

\colingfinalcopy % Uncomment this line for all SemEval submissions

\title{gundapusunil at SemEval-2020 Task 9: Syntactic Semantic LSTM Architecture for SENTIment Analysis of Code-MIXed Data }

\author{Sunil Gundapu \\
  Language Technologies Research Centre \\
  KCIS, IIIT Hyderabad \\
  Telangana, India \\
  {\tt sunil.g@research.iiit.ac.in} \\\And
  Radhika Mamidi \\
  Language Technologies Research Centre \\
  KCIS, IIIT Hyderabad \\
  Telangana, India \\
  {\tt radhika.mamidi@iiit.ac.in} \\}

\date{}

\begin{document}
\maketitle
\begin{abstract}
 The phenomenon of mixing the vocabulary and syntax of multiple languages within the same utterance is called Code-Mixing. This  is more evident in multilingual societies. In this paper, we have developed a system for SemEval 2020: Task 9 on Sentiment Analysis for Code-Mixed Social Media Text. Our system first generates two types of embeddings for the social media text. In those, the first one  is character level embeddings to encode the character level information and to handle the out-of-vocabulary entries and the second one is FastText word embeddings for capturing morphology and semantics. These two embeddings were passed to the LSTM network and the system outperformed the baseline model.
  
\end{abstract}

\blfootnote{
    %
    % for review submission
    %
    % % final paper: en-uk version 
    %
     \hspace{-0.65cm}  % space normally used by the marker
     This work is licensed under a Creative Commons 
     Attribution 4.0 International Licence.
     Licence details:
     \url{http://creativecommons.org/licenses/by/4.0/}.
    % 
    % % final paper: en-us version 
    %
    % \hspace{-0.65cm}  % space normally used by the marker
    % This work is licensed under a Creative Commons 
    % Attribution 4.0 International License.
    % License details:
    % \url{http://creativecommons.org/licenses/by/4.0/}.
}

\section{Introduction}

Code-Mixing is a phenomenon which is evident in multilingual societies (Shana Poplack et al., 2003). It reflects the use of distinct grammatical systems and vocabulary of the languages being used simultaneously in a single utterance or conversation. This technique used in communication commonly is widely found today in popular social media platforms like Twitter, Facebook, Instagram in the form of posts, comments, replies, especially in chats. This is evident in multilingual societies like India, Canada, Ireland, South Africa, Switzerland, and many others.

India has officially recognized 22 regional languages\footnote{https://en.wikipedia.org/wiki/Languages\_of\_India}. So, in multilingual societies like India, most of the social media users predominantly integrate the well-known language, like English, with their native languages. 560 million Internet users\footnote{https://www.internetworldstats.com/stats7.htm} in India exchange information by mixing their regional languages with prominent language like English, which produces a huge amount of code-mixed social media corpus. One such trending combination is the mixing of Hindi and English with the output in \textbf{Hinglish}\footnote{https://en.wikipedia.org/wiki/Hinglish} (Hi-En) code-mixed data. Consider the example sentence which illustrates the code-mixing phenomenon being addressed in this paper. 
“\textit{Congratulations/\textbf{Eng} Sir/\textbf{Eng} Ji/\textbf{Hin} Dobara/\textbf{Hin} PM/\textbf{Eng} banee/\textbf{Hin} ki/\textbf{Hin} hardik/\textbf{Hin} subhkamnaye/\textbf{Hin} aapko/\textbf{Hin} ./\textbf{O}}”. (Translation into English: Congratulations sir, Best wishes to become Prime Minister again). The words followed by language tags /Hin, /Eng, and /O correspond to Hindi, English and Other respectively.

Sentiment analysis of code-mixed data has become a prominent research area in recent times in the field of NLP. But identifying the sentiment in code-mixed data is hard since it poses the following challenges: (i) Romanized code-mixed data is noisy and ambiguous in nature. (ii) Accessible datasets are smaller in size to tune neural networks. (iii) Frequent occurence of non-standard spellings (such as pyaaarr, goooood). (iv) The phrase/word contractions (cmng for coming, IDK for I don’t know). (v) Spelling variations. A single word pyaar (love), can be written as “piyar”, “pyaarrrr”, “peyar”, “pyar”, or “piyaar”, etc. To handle these challenges we postulate that FastText embeddings enrich the word vectors with sub-word information and that character level embeddings should be able to assist deep learning models to handle unknown Hindi words.

In this paper, we propose models to predict the sentiment label of a given code-mixed tweet/text. The sentiment labels are positive, negative, or neutral and the code-mixed languages selected are Hindi and English. This task is conducted in \textbf{CodaLab}\footnote{https://competitions.codalab.org/competitions/20654} website and our CodaLab username is \textbf{gundapusunil}. All our models are trained using only the trained dataset provided by SentiMix organizers. We started experiments with traditional machine learning models like Logistic Regression, Support Vector Machines but the f1-score on development set was below 0.63 then we moved to complex models like Long Short Term Memory (LSTM) based models with different types of word embeddings where we were able to conquer the baseline model.

Our paper is divided into the following sections: We begin with an introduction to code-mixing and its challenges in Section 1. Related work of various code-mixing strategies demonstrated in Section 2. Section 3 present the description of SentiMix dataset. We then discuss the pre-processing steps and compare machine learning and deep learning approaches with baseline model results in Section 4. The results are reported in Section 5, and Section 6 concludes the paper.

\section{Related Work}

Several studies have been made in the areas of sentiment analysis and code-mixed data. One of the earlier studies on code-mixed data was proposed by Gold (1967) with the goal of language identification in which it was stated that the structure of the language is procured by learning the language structure from the given text and informant. Braj B., Kachru (1976) described the structure of multilingual languages and language dependency in linguistic convergence of code-mixing from an Indian Perspective. SentiWordNet for English language introduced by Esuli and Sebastiani (2006), became the primary source for all sense based lexical analysis and opinion classification.

Later, researchers have extended the work on machine learning and sentiment analysis methods for Indian languages. In their study Siersdorfer, Chelaru et al. (2010) used the SVM and Na\"ive Bayes classifiers to label millions of comments for sentiment polarity. R. Sharma and P. Bhattacharyya (2014) developed a lexicon-based sentiment analyzer for product reviews and the same subjective lexicon-based model has been extended to Punjabi language. For Malayalam movie reviews D. S. Nair, J. P. Jayan et al. (2014, 2015) initially came up with a rule based system for sentiment analysis, later improved the system with the support of the machine learning algorithms. MIKE 2015 sentiment analysis task for Hindi and Bengali used the Multinomial Na\"ive Bayes classifier with the features for building the vector space constrained by filtering the words based on WordNet. 

In recent years, researchers have seen a huge improvement in the task of sentiment analysis of English as well as Hinglish using deep neural networks. M.G. Jhanwar, A. Das (2018) proposed an ensemble of character-trigrams based LSTM model and a n-grams based Multinomial Na\"ive Bayes model to classify the sentiments of Hinglish code-mixed data. Shalini K, Barathi Ganesh HB et al. (2018) addressed the performance of distributed representation methods in sentiment analysis and reported comparisons among different machine learning and deep learning techniques. Other attempts include using sub-word level compositions with LSTMs to capture sentiment at the morpheme level (Joshi et al., 2016). We attempted to perform SemEval 2020 task-9 (Patwa et al., 2020) with various classification and deep learning models to analyze the results and also how such models contributed to a great advance in this task.

\section{Dataset}

In this paper, we used the dataset provided by Task 3: SentiMix in SemEval 2020. The corpus contains a total of 20000 tweets and it is sub-divided into three sets (train, validation and test). Each corpus except test contains code-mixed tweets along with their corresponding sentiment labels. These code-mixed tweets are tokenized into tokens. And the tokens of each tweet are separated by a new line. Each token is manually annotated with a language identification tag which are: Hindi (Hin), English (Eng), Other (O). 

\begin{table}[h!]
  \begin{center}
    \begin{tabular}{l|l|l|l|l}
      \hline
      \textbf{Dataset} & \textbf{Positive} & \textbf{Negative} & \textbf{Neutral} & \textbf{Total}\\
      \hline
      Train & 4634 & 4102 & 5264 & 14000 \\ % <--
      Valid & 982 & 890 & 1128 & 3000 \\ % <--
      Test & 1000 & 900 & 1100 & 3000 \\ % <--
      \hline
      &&&&20000\\
      \hline
      
    \end{tabular}
  \end{center}
  \caption{\label{font-table} Dataset Statistics. }
\end{table}

Table 1 shows the distribution of sentiment classes in the SentiMix dataset.  Table 2 shows some examples of code-mixed tweets from the SentiMix dataset. Here, the first column contains Hinglish tweet, the second column contains English translation of tweet, the third column contains sentiment label.

\begin{table}[h!]
    \centering
    \begin{adjustbox}{width=\textwidth}
    \begin{tabular}{l|l|l}
      \hline
      \textbf{Code-Mixed Tweet} & \textbf{Approximate English Translation} & \textbf{Sentiment Label}\\
      \hline
      All the best Team India Jeet ke aana & Team India all the best, come back with win. & Positive \\ % <--
      Aap bhi aisa drama kr do kam se km & You also do this drama and at least. & Negative  \\ % <--
      Aisa PM naa hua hai aur naa hee hoga & Neither there has been a PM like him, nor there will be & Positive \\ % <--
      \hline
    \end{tabular}
    \end{adjustbox}
  \caption{\label{font-table} Example Code-Mixed Tweets. }
\end{table}

\section{System Architecture}

In this section, we present our models that are trained and validated on the SentiMix dataset described in the previous section. We compare our approach with Machine Learning and Neural Network based baselines. The full code of system architecture can be found at
GitHub\footnote{https://github.com/SunilGundapu/SENTIment-Analysis-on-Hindi-English-Code-MIXed-Data}.

\subsection{Preprocessing of Code-Mixed tweets}

Our code-mixed data consists of excessive noise in the form of punctuations, Uniform Resource Locator (URL's), few Devanagari script Hindi words, twitter mentions, hashtags, stop words, etc. In the preprocessing step, we take a stab to overcome the noise in the the data by remove/normalize the unnecessary tokens. Figure 1 explains the code-mixed dataset pre-processing pipeline. The input for the pipeline is a tokenized tweet and the output is a cleaned tweet.

\begin{center}
  \begin{figure}[h!]
  \makebox[\textwidth]{\includegraphics[width=\textwidth, height = 4.5cm]{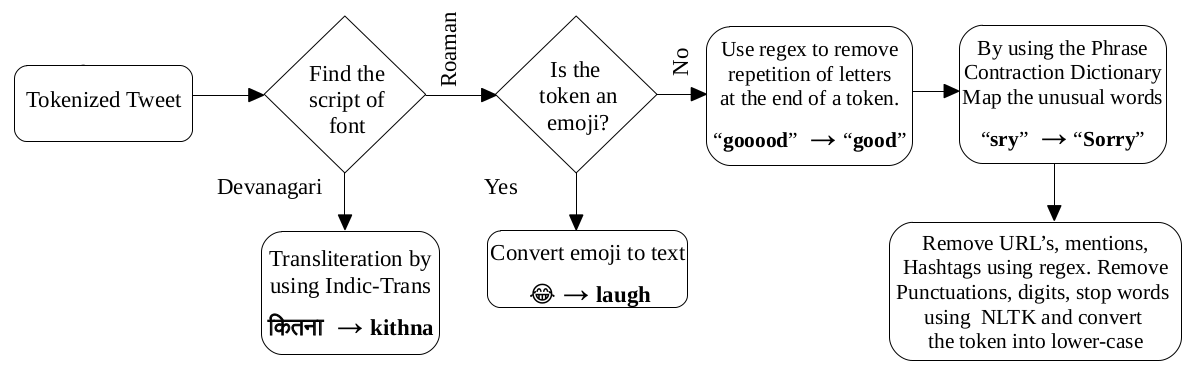}}
  \caption{Code-Mixed text preprocessing pipeline.}
  \end{figure}
\end{center}

\subsection{Classical Supervised Machine Learning Algorithms}

To design the finest system for sentiment analysis in code-mixed data, we begin our experiments with traditional machine learning algorithms like Support Vector Machines (SVM), One-vs-rest classifier with Logistic Regression (OvRLR), Random Forest Classifier (RFC) and  Multilayer Perceptron (MLP). The input for these methods is a single $\textit{\textbf{d}}$ dimensional feature vector of a single code-mixed tweet. 

We analyze the results of the above classical algorithms with the combination of two types of vectors. (i) Word level term frequency-inverse document frequency (tf -idf) vector and (ii) Glove word embeddings. For all tokens in the code-mixed tweet a feature vector is created by averaging over $\textit{\textbf{d}}$ dimensional Glove embeddings and also it is experimented with tf-idf weighted averaging. The code-mixed tweet vector construction scheme is described below:

\begin{align*}
featureVector_{tweet} &= \dfrac{\sum_{n=1}^{N} \textbf{tf-idf}(token_i) \times  \textbf{Glove}(token_i)}{total\,no.\,of\,tokens\,in\,tweet\,(N)}
\end{align*}

Empirically, we found that standard averaging of Glove and tf-idf gave better results than normal tf-idf weighted averaging.

\subsection{Deep Neural Networks}

In this subsection, we describe the character and word embedding based deep neural network called \say{Syntactic and Semantic LSTM (SS-LSTM)} that gave better predictions on our corpus. Initially, We tried with Word2Vec (T. Mikolov et al., 2013), GloVe (J. Pennington et al., 2014), FastText (Bojanowski et al., 2016), Character embeddings for each word in the input code-mixed tweet. We train a simple LSTM model using each of these embeddings to test the effectiveness of these embeddings for sentiment classification. FastText and Character level embeddings gave slightly better results than other embeddings. By considering these results we modeled the SS-LSTM architecture given below. 

\subsubsection{Character Level Embeddings } 

Character level embeddings use a one-dimensional convolutional neural network (1D-CNN) to find the numeric representation of words by looking at their character-level compositions. 1D-CNN (Yoon Kim, 2014) is an algorithm capable of handling unseen words and also extracting syntactic information from the segments of input. The character embedding step converts tweet tokens into a $\textit{\textbf{d}} \times \textit{\textbf{T}}$ matrix. \textit{\textbf{d}} is the dimension of vector and \textit{\textbf{T}} is the number of tokens in code-mixed tweet.

\subsubsection{Word Level Embeddings}

For word level embeddings we used the Facebook’s FastText. The main advantage of FastText embeddings is to capture the hidden knowledge about a language, like word analogies or semantic. And it is looking into the internal structure of words, which could be very useful for morphologically rich languages like Hindi. The  FastText enrich word vectors with subword information.

\subsubsection{Model Architecture}

\begin{center}
  \begin{figure}[h!]
  \makebox[\textwidth]{\includegraphics[width=13.5cm,height=6.5cm]{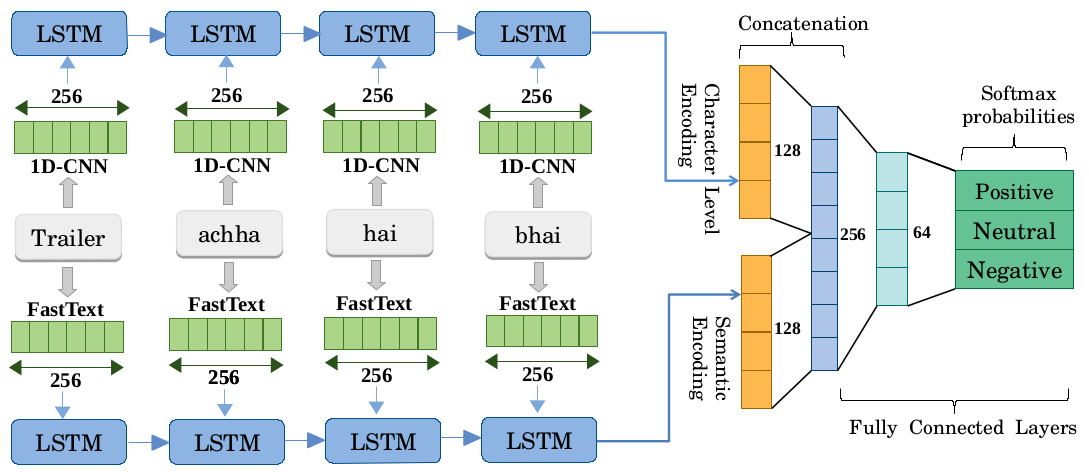}}
  \caption{System (SS-LSTM) Architecture.}
  \end{figure}
\end{center}

We model the task of SentiMix as a multi-class classification problem where given a code-mixed tweet, the model outputs probabilities of it belonging to three output classes - Positive, Negative, and Neutral. The proposed system architecture (SS-LSTM) is shown in Figure 2. The input tweet is fed into 1D-CNN and FastText. These two word embedding models generate two $\textit{\textbf{d}} \times \textit{\textbf{T}}$ matrices, one is for the 1D-CNN and the other for the FastText. Here the dimension of each matrix is $\textit{\textbf{256}} \times \textit{\textbf{T}}$. These two word embedding matrices are passed to two LSTM layers.  One LSTM layer uses a character level embeddings, whereas the other layer uses a  FastText word embeddings. These two layers learn syntactic and semantic feature representation and encode sequential patterns in the tweet. And each LSTM layer gives a  $\textit{\textbf{128}} \times \textit{\textbf{1}}$ dimensions vector.

The character embedding LSTM layer output is  $C\in{R^{128\times1}}$ and the FastText word embedding LSTM layer output is $W\in{R^{128\times1}}$.
These two output feature representations are row-wise concatenated and the output vector dimension is $O\in{R^{256\times1}}$. The output vector $\textit{\textbf{O}}$ passed to a fully connected network with one hidden layer which models interactions between these features and
outputs probabilities per sentiment class. We used \textbf{Keras} neural network library to implement this model. $O\in{R^{256\times1}}$

\section{Results}
A summary of results from various techniques on the SentiMix test dataset is present in Table 3. SS-LSTM gave the best performance on f1-score for each sentiment class as well as on average f1-score. Our results thus indicate that combining syntactic and semantic representations in SS-LSTM outperforms individual LSTM-Character and LSTM-FastText embedding models.

\begin{table}[h!]
  \begin{center}
    \begin{tabular}{ccccc}
        \hline
        \textbf{Model} & \textbf{Representations} & \textbf{Precision} & \textbf{Recall} & \textbf{f1-Score} \\
        \hline
        SVM & TF-IDF avg  & 0.6315    & 0.6373 & 0.6308     \\
        RFC & TF-IDF avg  & 0.6260     & 0.6323  & 0.6261     \\
        OvRLR & TF-IDF avg & 0.6391     & 0.6426  & 0.6377    \\
        MLP & TF-IDF avg & 0.6505      & 0.6433  & 0.6454    \\
        \hline
        SVM & TF-IDF and Glove avg  & 0.6409    & 0.6384 & 0.6412     \\
        RFC & TF-IDF and Glove avg  & 0.6311     & 0.6358  & 0.6356     \\
        OvRLR & TF-IDF and Glove avg & 0.6357     & 0.6393  & 0.6343    \\
        MLP & TF-IDF and Glove avg & 0.6485      & 0.6523  & 0.6445    \\
        \hline
        LSTM & Word2Vec & 0.6565 & 0.6538 & 0.6523 \\
        LSTM & FastText & 0.6605 & 0.6533 & 0.6554 \\
        LSTM & Glove & 0.6585 & 0.6523 & 0.6545 \\
        LSTM & 1D-CNN & \textbf{0.6825} & 0.6610 & 0.6651 \\
        LSTM & BERT & 0.6800 & 0.6712 & 0.6742 \\
        \hline
        SS-LSTM & FastText and 1D-CNN & 0.6819 & \textbf{0.6773} & \textbf{0.6758} \\
    \hline
    \end{tabular}
  \end{center}
  \caption{\label{font-table} Results on test data for Hindi-English.}
\end{table}

Machine learning models with tf-idf feature representations gave the approximate baseline results. We observe that the tf-idf weighted average of GloVe performed better than the simple average of vectors. And we used the grid search (Srivastava et al., 2014) to find the better hyper-parameters like number of LSTM layers, learning rate, and the number of epochs. We used GPU for training deep learning models.

\section{Conclusion}
In this paper, we experimented the code-mixed dataset with various machine learning and deep learning models. We see that the LSTM models performed far better than traditional ML methods. In the first phase of the SentiMix competition (development set), we were able to achieve a score of \textbf{0.6357}. But in the second phase (test dataset), our best score was only \textbf{0.6758}. After competition we attain the f1-score of \textbf{0.6789} by changing few parameters like learning rate and number of LSTM's. Till now we handled problems like unseen words, spelling variations, dataset imbalance, emojis, short form of words, etc. In future work, we plan to focus on issues like free ordering of words in sentence constructions, short sentences with unclear semantic structure, etc. And we would like to explore more deep neural network architectures that can capture sentiments in code-mixed data.


\begin{thebibliography}{}

\bibitem[\protect\citename{}]{}
Parth Patwa, Gustavo Aguilar, Sudipta Kar, Suraj Pandey, Srinivas PYKL, Bj{\"o}rn Gamb{\"a}ck, Tanmoy Chakraborty, Thamar Solorio, and Amitava Das. 
\newblock 2020.
\newblock Semeval-2020 task 9: Overview of sentiment analysis of code-mixed tweets.
\newblock In \emph{Proceedings of the 14th International Workshop on Semantic Evaluation ({S}em{E}val-2020)}, Barcelona, Spain, December.
\newblock Association for Computational Linguistics.

\bibitem[\protect\citename{Borschinger and Johnson}2011]{borsch2011}
Poplack, Shana and Walker, James.
\newblock{ 2003.}
\newblock{Pieter Muysken, Bilingual speech: a typology of code-mixing.}
\newblock{Cambridge: Cambridge University Press, Pp. xvi+306.}
\newblock{Journal of Linguistics. 39. 678 - 683. 10.1017/S0022226703272297.}

\bibitem[\protect\citename{Gold and TR}1967]{Gold:72}
E. M. Gold and T. R. Corporation.
\newblock 1967.
\newblock {\em Language identification in the limit}, Inf. Control, vol~10, no~5, pp~447–474.

\bibitem[\protect\citename{Gold and TR}2006]{Esuli:06}
Braj B. Kachru.
\newblock 1978.
\newblock{Toward Structuring Code-Mixing:
An Indian Perspective.}
\newblock {\em International Journal of the Sociology of Language}, 16:27-46.

\bibitem[\protect\citename{Gold and TR}2006]{Esuli:06}
A. Esuli, F. Sebastiani, and V. G. Moruzzi.
\newblock 2006.
\newblock {\em SENTIWORDNET: A Publicly Available Lexical Resource for Opinion Mining}, Proc. Lr, vol~0, pp~417–422.


\bibitem[\protect\citename{Borschinger and Johnson}2011]{borsch2011}
S. Siersdorfer, S. Chelaru, W. Nejdl, and J. San Pedro.
\newblock 2011.
\newblock How useful are your comments?.
\newblock In \emph{Proceedings of 19th International Conference on World Wide Web 2011}, vol~15, pp~891–900.

\bibitem[\protect\citename{Borschinger and Johnson}2011]{borsch2011}
R. Sharma and P. Bhattacharyya.
\newblock 2014.
\newblock A sentiment analyzer for hindi using hindi senti lexicon.
\newblock In \emph{Proceedings of 11th International Conference on Natural Language Processing 2014}, p~150.

\bibitem[\protect\citename{Borschinger and Johnson}2011]{borsch2011}
K. Shalini, H. B. Ganesh, M. A. Kumar and K. P. Soman.
\newblock 2018.
\newblock Sentiment Analysis for Code-Mixed Indian Social Media Text With Distributed Representation.
\newblock \emph{International Conference on Advances in Computing, Communications and Informatics (ICACCI)}, Bangalore, pp~1126-1131.

\bibitem[\protect\citename{}]{}
D. S. Nair, J. P. Jayan, E. Sherly.
\newblock 2015.
\newblock Sentima-sentiment extraction
for malayalam.
\newblock In \emph{ Advances in Computing, Communications and Informatics (ICACCI), 2015 International Conference on. IEEE}, pp~2381-2384.

\bibitem[\protect\citename{}]{}
K. Sarkar and S. Chakraborty.
\newblock 2015.
\newblock A sentiment analysis system for indian language tweets.
\newblock In \emph{International Conference on Mining Intelligence and Knowledge Exploration.}, Springer, pp~694-702.

\bibitem[\protect\citename{}]{}
Madan G. Jhanwar and Arpita Das.
\newblock 2018.
\newblock  \emph{An Ensemble Model for Sentiment Analysis of Hindi-English Code-Mixed Data}, CoRR, vol~abs/1806.04450.

\bibitem[\protect\citename{}]{}
A. Joshi, A. Prabhu, M. Shrivastava, V. Varma.
\newblock 2016.
\newblock Towards sub-word level compositions for sentiment analysis of hindi english code mixed text.
\newblock In \emph{Proceedings of COLING 2016, the 26th International Conference on Computational Linguistics: Technical Papers}, pp~2482-2491.

\bibitem[\protect\citename{}]{}
Patra, Braja and Das, Dipankar and Das, Amitava.
\newblock 2018.
\newblock Sentiment Analysis of Code-Mixed Indian Languages: An Overview of SAIL Code-Mixed Shared Task @ICON-2017.

\bibitem[\protect\citename{}]{}
Gupta, Umang and Chatterjee, Ankush and Srikanth, Radhakrishnan and Agrawal, Puneet.
\newblock 2017.
\newblock A Sentiment-and-Semantics-Based Approach for Emotion Detection in Textual Conversations.

\bibitem[\protect\citename{}]{}
A. Joulin, E. Grave, P. Bojanowski, and T. Mikolov.
\newblock 2016.
\newblock Bag of tricks for efficient text
classification.
\newblock In \emph{arXiv preprint arXiv:1607.01759}. 

\bibitem[\protect\citename{}]{}
T. Mikolov, I. Sutskever, K. Chen, G. S. Corrado, and J. Dean.
\newblock 2013.
\newblock Distributed representations of words and phrases and their compositionality.
\newblock In \emph{Advances in neural information processing systems}, pp~3111-3119.

\bibitem[\protect\citename{}]{}
J. Pennington, R. Socher, and C. D. Manning.
\newblock 2014.
\newblock Glove: Global vectors for word
representation.
\newblock In \emph{EMNLP}, vol~14, pp~1532–1543.

\bibitem[\protect\citename{}]{}
A. Joshi, A. Prabhu, M. Shrivastava, V. Varma.
\newblock 2014.
\newblock Convolutional Neural Networks for Sentence Classification.
\newblock In \emph{Proceedings of the 2014 Conference on Empirical Methods in Natural Language Processing (EMNLP)}, pp~1746-1751.

\bibitem[\protect\citename{}]{}
Nitish Srivastava, Geoffrey Hinton, Alex Krizhevsky,
Ilya Sutskever, and Ruslan Salakhutdinov.
\newblock 2014.
\newblock \emph{Dropout: a simple way to prevent neural networks from overfitting}, 15(1):1929–1958.

\end{thebibliography}
\end{document}